\documentclass{article} 
\usepackage{geometry}
\usepackage{amssymb}
\usepackage{graphicx}
\usepackage[figuresright]{rotating}
\usepackage{tabularx}
\usepackage{color}
\usepackage{soul}
\usepackage{epstopdf}
\usepackage[utf8]{inputenc}
\usepackage[]{siunitx}
\usepackage{hyperref}
\usepackage{xstring}
\usepackage{cleveref}
\usepackage[authoryear, sort]{natbib}
\usepackage{caption}
\usepackage{booktabs}

\DeclareCaptionLabelFormat{andtable}{#1~#2  \&  \tablename~\thetable}
\begin{document}

\title{Sequence to sequence pretraining for a less-resourced Slovenian language}

\author{Matej Ul\v{c}ar, Marko Robnik-\v{S}ikonja \\
University of Ljubljana, Faculty of Computer and Information Science \\
Ve\v{c}na pot 113, Ljubljana, Slovenia \\
\{matej.ulcar, marko.robnik\}@fri.uni-lj.si
}

\date{}

\maketitle

\begin{abstract}

Large pretrained language models have recently conquered the area of natural language processing. 
As an alternative to predominant masked language modelling introduced in BERT, the T5 model has introduced a more general training objective, namely sequence to sequence transformation, which includes masked language model but more naturally fits text generation tasks such as machine translation, summarization, question answering, text simplification, dialogue systems, etc. The monolingual variants of T5 models have been limited to well-resourced languages, while the massively multilingual T5 model supports 101 languages. In contrast, we trained two different sized T5-type sequence to sequence models for morphologically rich Slovene language with much less resources and analyzed their behavior on 11 tasks. 
Concerning classification tasks, the SloT5 models mostly lag behind the monolingual Slovene SloBERTa model but are useful for the generative tasks.

 \textbf{Keywords:} \textit{natural language processing, pretrained language models, sequence to sequence models, transformers, T5 model, low-resource languages, Slovene} 
\end{abstract}

\section{Introduction}
Recent state-of-the-art natural language processing (NLP) solutions are based on the transformer neural network architecture \citep{Vaswani2017}. The main research direction is to produce (ever larger) pretrained language models (PLMs) with billions of parameters with the objective to contain as much human knowledge as possible \citep{bommasani2021opportunities}. Such models require large amounts of training data and are computationally expensive to train. Most very large models have been trained for English and a few high-resource languages, such as Chinese, French, or German. Massively multilingual models, trained on around 100 languages, have also been released, but their performance lags behind their monolingual and few-lingual equivalents \citep{ulcar2021evaluation}.
For some of these 100 less-resourced languages, there is a growing number of smaller models (though still in the range of a few 100 million parameters) trained on the BERT \citep{Devlin2019} or RoBERTa \citep{liu2019roberta} architecture.

BERT \citep{Devlin2019} is a masked language model, utilizing the encoder stack of the transformer architecture to capture the semantic representation of the input text. This makes it very suitable for solving classification tasks. Another popular type of large language models are from the GPT family, such as GPT-2 \citep{Radford2019GPT2} and GPT3 \citep{Brown2020GPT3}, which are generative models and utilize only the decoder stack of the transformer. In contrast to these models, sequence to sequence (seq2seq) models such as BART \citep{lewis-etal-2020-bart} and T5 \citep{raffel2020exploring} utilize both encoder and decoder stack of the transformer. Such models can treat every problem as a text-to-text transformation and solve it similarly, without adapting the training procedure for each task.

In this work, we present two new sequence to sequence models for the less-resourced Slovene language based on the T5 architecture and its training tasks. We aim to analyze the amount of required data for such models to be effective and the role the richer morphology plays for seq2seq models. Namely, while English has a fixed word order and relatively few word forms for each word, this is not the case for most other languages. This might not be problematic in text classification, which is a typical task for large pretrained models, while text generation tasks are more challenging for morphologically-rich languages. We qualitatively and quantitatively test Slovene T5 models on three text generation tasks: lemmatization, summarization, and text simplification.
We believe our results might be indicative for other less-resourced languages in terms of datasets, training, and expected results.

The work is split into further four sections. In \Cref{sec:relatedwork}, we summarize the related work and briefly describe the T5 architecture In \Cref{sec:SloT5}, we present the architecture and training of the Slovene T5 models, which are evaluated in \Cref{sec:evaluation}. We discuss the findings and their implications in \Cref{sec:discussion}.

\section{Related work}
\label{sec:relatedwork}

T5 model \citep{raffel2020exploring} is an encoder-decoder transformer, trained on several supervised and self-supervised pretraining tasks. The supervised tasks used were the tasks from the GLUE \citep{wang-etal-2018-glue} and SuperGLUE \citep{wang2019superglue} benchmarks, as well as translation and summarization tasks. The self-supervised task used was the span-corruption task. 
In this task, randomly selected token spans are replaced with a special mask token (a sentinel token). The goal of the task is to generate the masked spans. During pretraining, 15\% of tokens were masked in spans with an average length of 3 tokens. The encoder stack receives the tokenized input text. The self-attention mechanism attends to the whole input in the encoder. The output of the encoder is then fed into the decoder stack, which generates the target text. A causal mask is used to prevent the self-attention mechanism in the decoder to attend to the "future" output. At each timestep, the model "sees" the whole input sequence and the part of the output sequence generated at previous timesteps.
Several T5 models for English have been trained and released. They differ in size, ranging from 60 million to 11 billion parameters.

\cite{xue2021mt5} have trained multilingual T5 models (mT5) of various sizes. The mT5 models were trained on a large multilingual mC4 corpus, containing 101 languages, and a total of $6.3\cdot10^{12}$ tokens. The Slovenian portion of the corpus contains 8.8 billion tokens. The mT5 models were trained simultaneously on all 101 languages on the span corruption task only. 

BART \citep{lewis-etal-2020-bart} is another popular encoder-decoder transformer model. The main difference between BART and T5 is in the choice of the pretraining tasks. Similarly to T5 and mT5, BART was trained on the span corruption task. Additionally, token deletion, sentence permutation, and document rotation tasks were used during pretraining. \cite{liu-etal-2020-multilingual} trained a multilingual BART (mBART) model on 25 languages, using the span corruption (masking 35\% of the words) and sentence permutation training tasks. \cite{tang-etal-2021-multilingual} extended the existing mBART model to further 25 languages, thus covering 50 languages, including Slovene.

Several monolingual models, based on the T5 architecture, have been released for high-resource languages other than English, such as Chinese Mengzi \citep{zhangetal2021mengzi}, Arabic AraT5 \citep{nagoudietal2021arat5}, and Italian IT5 \citep{sarti-nissim2022it5}. While \cite{nagoudietal2021arat5} observe the improvement of AraT5 over mT5 across all evaluation tasks, \cite{sarti-nissim2022it5} note that especially for summarization IT5 lags behind the benchmark models. \cite{sarti-nissim2022it5} also observed that scaling the model size does not uniformly correspond to improvements in performance. On average, the small IT5 improves the most over the comparable mT5 model, while larger IT5 models show much smaller or no improvements over comparable mT5 models and, in some cases, perform even worse than the small IT5 model.

The presented Slovene SloT5 models partially confirm and partially contradict the above findings. On one hand, we use much more challenging text classification tasks (Slovene translation of the SuperGLUE benchmark suite); therefore, the classification performance of SloT5 models consistently lags behind the BERT-like monolingual Slovene models. On the other hand, while the small SloT5 model is successful for text generation tasks, the amount of training data and training time might not be sufficient to make the large SloT5 model really competitive in the text generation tasks.

\section{Slovene T5 models}
\label{sec:SloT5}
In this section, we present the newly created Slovene T5 models (named SloT5). First, we describe the training data, followed by the description of architecture and training.

\subsection{Training data}
We trained Slovene SloT5 models on large Slovene corpora, covering a wide spectrum of genres, from fiction books to newspapers, academic language, internet slang, etc. We included Gigafida, Janes, KAS, SiParl, and SlWaC corpora. The corpora details are given below and summarized in Table~\ref{tab:corpora}.

Gigafida 2.0 \citep{krek-etal-2020-gigafida} is a general standard language corpus composed of fiction and non-fiction books, newspapers, school textbooks, texts from the internet, etc. The Janes corpus \citep{fivser2016janes} is a corpus of non-standard language composed of several subcorpora. Each subcorpus contains texts from a certain social medium or a group of similar media, including Twitter, blog posts, forum conversations, user comments on news site articles, etc. We used all Janes subcorpora, except Janes-tweet, since the contents of that subcorpus are encoded and need to be individually downloaded from Twitter, which is a lengthy process as Twitter limits the access speed. KAS (Corpus of Academic Slovene) \citep{erjavec2021kas} consists of PhD, MSc, MA, BSc, and BA theses written in Slovene between 2000 and 2018. SiParl \citep{pancur2020siparl} contains minutes of Slovene national assembly between 1990 and 2018. SlWaC \citep{ljubevsic2011hrwac} is a web corpus collected from the Slovene top-level web domain .si. 

\begin{table}[htb]
    \caption{Corpora used in training of SloT5 models with their sizes in billion of tokens and words. Janes subcorpora used are listed separately, but we show their combined size.}
    \label{tab:corpora}
    \begin{tabular}{llcc}
    Corpus & Genre & Tokens & Words \\
    \midrule
    Gigafida 2.0 \citep{gigafida} & general language & $1.33$  & $1.11$ \\
    KAS \citep{11356/1448} & academic & $1.70$  & $1.33$ \\
    siParl 2.0 \citep{11356/1300} & parliamentary & $0.24$  & $0.20$ \\
    slWaC 2.1  \citep{ljubevsic2011hrwac} & web crawl & $0.90$  & $0.75$ \\ 
    Janes \citep{fivser2016janes} & social media  & $0.10$  & $0.08$ \\
    -- Janes-Wiki \citep{11356/1137} & \multicolumn{3}{l}{-- Wikipedia talk pages} \\
    -- Janes-Blog \citep{11356/1138} & \multicolumn{3}{l}{-- Slovene blogs} \\
    -- Janes-Forum \citep{11356/1139} & \multicolumn{3}{l}{-- Slovene forums} \\
    -- Janes-News \citep{11356/1140} & \multicolumn{3}{l}{-- comments on online news articles}  \\
    \midrule
    \multicolumn{2}{l}{Total} & $4.27$ & $3.47$ \\ 
    \multicolumn{2}{l}{Total after deduplication} & $4.20$ & $3.41$ \\ 
    \end{tabular}
\end{table}

We deduplicated the corpora, using the Onion tool \citep{pomikalek2011removing}. After the deduplication, the training dataset contained about 4 billion tokens. Finally, before training the models, we encoded the text into subword byte-pair-encodings using a sentencepiece\footnote{\href{https://github.com/google/sentencepiece}{https://github.com/google/sentencepiece}} model. We used the sentencepiece model that was trained for SloBERTa \citep{ulvcar2021sloberta} and contains 32,000 subword tokens in its vocabulary.

\subsection{Architecture and training of SloT5}
We trained Slovene T5 models of two different sizes: T5-sl-small and T5-sl-large. The smaller model has 8 encoder and 8 decoder layers, in total, about 60 million parameters. The larger model has 24 encoder and 24 decoder layers, in total, about 750 million parameters. All the models were trained in the same manner, i.e. on the same tasks with the same amount of data and the same optimizer. We compare two smaller models, which differ in the amount of training (1 or 5 epochs), and three larger models (1, 3, or 5 epochs).

We trained the models on a mixture of two self-supervised pre-training tasks: i.i.d. (independent and identically distributed) denoising and span corruption, suggested by \cite{raffel2020exploring}. In the i.i.d. denoising task, 15\% tokens were randomly corrupted, i.e. replaced by a sentinel token. Each token has an equal probability of being corrupted (identically distributed) and all the corruption/replacing events are independent from each other. The goal of the task is to denoise the sequence by generating the correct token in place of the sentinel. This task is identical to the span corruption task, described in \Cref{sec:relatedwork}, except that all spans have the length of one token. The span corruption task used in training SloT5 is identical to the one used for training English T5 and multilingual mT5 models, with 15\% of tokens corrupted and an average corrupted span length of 3 tokens.

The T5-sl-small\textsubscript{1} and T5-sl-large\textsubscript{1} models were trained for 1 million steps, with a batch size of 4096 tokens, in total a bit less than 1 epoch. This amount of training is supposed to be sufficient, considering the ratio between the training tokens and the number of the model parameters \citep{1epoch}. Additionally, we trained T5-sl-small\textsubscript{5} for 763,000 steps, with a batch size of 32,768 tokens, in total around 5 epochs. T5-sl-large\textsubscript{3} and T5-sl-large\textsubscript{5} were trained with a batch size of 8192 tokens, for 1.83 million steps and 3.05 million steps, respectively, which results in about 3 and 5 epochs.
We trained the models on a DGX A-100 machine, using four 40 GB A100 GPUs. The training took about 3 days for T5-sl-small\textsubscript{1}, about 12 days for T5-sl-small\textsubscript{5}, about 3 weeks for T5-sl-large\textsubscript{1}, about 4 weeks for T5-sl-large\textsubscript{3}, and about 7 weeks for T5-sl-large\textsubscript{5}.

\section{Evaluation}
\label{sec:evaluation}
We evaluated our newly trained SloT5 models on 11 classification and generative tasks: named entity recognition, sentiment classification, lemmatization, text simplification, two summarization tasks on different datasets, and five (essentially classification) tasks from the Slovene SuperGLUE \citep{zagar2022slovene} benchmark (two question answering, two natural language inference, and a coreference resolution task). 

For classification tasks, we could use only the encoder stack of the T5 and added appropriate task-specific output heads on top of it, thus completely ignoring/bypassing the decoder stack. However, we decided to remodel the classification tasks into generative tasks, mimicking the evaluation procedure proposed by \citet{raffel2020exploring}. Therefore, each example contained only an input string and an output string.

Next, in \Cref{sec:tasks}, we describe all eleven evaluation tasks and explain their preprocessing for the seq2seq models. The details of fine-tuning the SloT5 models and other compared transformer models are contained in \Cref{sec:finetuning}. In \Cref{sec:results}, we present the results. We present qualitative analysis of the results on two tasks in \Cref{sec:analysis}.

\subsection{Evaluation tasks}
\label{sec:tasks}
In this section, we describe the evaluation tasks and their preprocessing for T5 models. For the named entity recognition (NER) task and SuperGLUE tasks, we show the examples of original entries and entries preprocessed for T5 modelling in Table~\ref{tab:t5encodings}. We did not apply any special preprocessing for the sentiment analysis classification task and the generative tasks.

\begin{table}[h!tb]
    \caption{Original examples and T5 formatted versions for each of the SuperGLUE tasks and the NER task. T5 formatted examples are in the CSV format where the first column is the input and the second the output.}
    \label{tab:t5encodings}
    \begin{tabular}{p{0.14\textwidth}p{0.86\textwidth}}
    \midrule
    BoolQ (original) & \{"label": true, "passage": "Kalcijev karbid - Kalcijev karbid je kemična spojina s kemično formulo CaC. Njegova glavna uporaba v industriji je pri proizvodnji acetilena in kalcijevega cianamida.", "question": "kalcijev karbid cac2 je surovina za proizvodnjo acetilena"\} \\
    BoolQ (T5~formatted) & "Sestavek: Kalcijev karbid - Kalcijev karbid je kemična spojina s kemično formulo CaC. Njegova glavna uporaba v industriji je pri proizvodnji acetilena in kalcijevega cianamida. Vprašanje: kalcijev karbid cac2 je surovina za proizvodnjo acetilena", "Pravilno." \\ 
    \midrule
    CB (original) & \{"premise": "Bil je kompleksen jezik. Ne zapisano, ampak predano. Lahko bi rekli, da je bil olupljen.", "hypothesis": "jezik je bil olupljen", "label": "entailment"\} \\
    CB (T5~formatted) & "premisa: Bil je kompleksen jezik. Ne zapisano, ampak predano. Lahko bi rekli, da je bil olupljen. hipoteza: jezik je bil olupljen", "implikacija" \\
    \midrule
    COPA (original) & \{"premise": "Moje telo je metalo senco na travo.", "choice1": "Sonce je vzhajalo.", "choice2": "Trava je bila pokošena.", "question": "cause", "label" :0\} \\
    COPA (T5~formatted) & "Premisa: Moje telo je metalo senco na travo. Prva možnost: Sonce je vzhajalo. Druga možnost: Trava je bila pokošena. Kaj je vzrok?", "prva" \\
    \midrule
    RTE (original) & \{"premise": "V Iraku še ni bilo najdenega orožja za množično uničevanje.", "hypothesis": "V Iraku najdeno orožje za množično uničevanje.", "label": "not\_entailment"\} \\
    RTE (T5~formatted) & "premisa: V Iraku še ni bilo najdenega orožja za množično uničevanje. hipoteza: V Iraku najdeno orožje za množično uničevanje.", "ni implikacija" \\
    \midrule
    WSC (original) & \{"target": \{"span1\_text": "skodelico", "span2\_text": "bila", "span1\_index": 4, "span2\_index": 9\}, "text": "Iz steklenice sem v skodelico nalival vodo, dokler ni bila polna.", "label": true\} \\
    WSC (T5~formatted) & "WSC: Iz steklenice sem v * skodelico * nalival vodo, dokler ni \# bila \# polna.", "Pravilno." \\
    \midrule
    NER  & 1130	4167	Bolj	O \\
    (original)       & 1130	4167	teoretično	O \\
                   & 1130	4167	pa	O \\
                   & 1130	4167	se	O \\
                    & 1130	4167	je	O \\
                    & 1130	4167	problema	O \\
                     & 1130	4167	lotil	O \\
                     & 1130	4167	Radical	B-ORG \\
                     & 1130	4167	Science	I-ORG \\
                     & 1130	4167	Journal	I-ORG \\
                     & 1130	4167	v	O \\
                     & 1130	4167	Londonu	B-LOC \\
                     & 1130	4167	.	O \\
    NER (T5~formatted) & "organizacije: Bolj teoretično pa se je problema lotil Radical Science Journal v Londonu .", "Radical Science Journal" \\
                     & "lokacije: Bolj teoretično pa se je problema lotil Radical Science Journal v Londonu .", "Londonu" \\
                     & "osebe: Bolj teoretično pa se je problema lotil Radical Science Journal v Londonu .", "brez" \\

\\    \bottomrule
    \end{tabular}
\end{table}

\subsubsection{Classification tasks}
Named entity recognition (NER) is a token-classification task, where each token is labelled as a named entity (NE) or not, and, if yes with the category of the NE. We used a dataset based on the ssj500k corpus v2.2 \citep{ssj500k}. We covered three categories of NEs: persons, locations, and organizations. To our knowledge, there is no standardized way of solving the NER task using seq2seq models. We first attempted to generate labels for each token in a sentence, but the dataset was overwhelmed by label "O", which covers all tokens that are not NEs and includes other named entity categories (e.g., products). We propose to solve the problem as a NE retrieval task. We prefixed each training sentence with each NE category, thus generating three times the number of training examples. See an example of the input and output in \Cref{tab:t5encodings}. 

The desired output is a comma-separated list of NEs in the sentence pertaining to the prefixed category.
*-If there are no NEs of the given category in a sentence, we set the output in the training set to the Slovene word "brez", meaning "none"/"empty". The resulting dataset still has most examples with the output "brez". We balanced the training dataset by omitting examples without NEs with 95\% probability. We followed the same procedure for the validation dataset, omitting 50\% of examples without NEs. However, the test set was not modified and we kept all such examples in it. 

Sentiment analysis (SA) is a sentence-level classification task composed of tweets, each labelled with one of three classes: "positive", "negative", or "neutral". We used Slovenian tweets from the Twitter sentiment dataset for 15 European languages \citep{mozetivc2016multilingual}. Each class label was translated into Slovene as a single word to be generated by the model; no other formatting changes were needed.

Slovene SuperGLUE \citep{zagar2022slovene} benchmark was translated from the English SuperGLUE benchmark \citep{wang2019superglue}. It contains two separate datasets: one was translated using machine translation and the other by human translators. Human translated datasets are of higher quality than machine translations but smaller in size for most tasks, as only subsets of the datasets were translated. We used five tasks from the SuperGLUE benchmark: Boolean question answering (BoolQ), Choice of Plausible Alternatives (COPA), CommitmentBank (CB), Recognizing Textual Entailment (RTE), and The Winograd Schema Challenge (WSC). For BoolQ, CB, COPA, and RTE, we used larger machine-translated datasets. For the WSC task, we used the human translated dataset (WSC is impossible to translate with machine translation tools).

BoolQ consists of triples: a passage, a question based on the passage, and an answer (true or false). In the COPA task, the goal is to pick the correct of the two given sentences, which correctly relates to the given premise and relation (cause or effect). CB and RTE datasets contain textual entailment tasks, where given a premise and a hypothesis, the goal is to predict whether the hypothesis entails the premise or not. In the WSC task, two spans in a short text are highlighted. The goal is to identify, using world knowledge and commonsense reasoning, whether both highlighted spans refer to the same entity. 

SuperGLUE tasks have multiple attributes. As we can only feed a single string input to the T5 model, we have prefixed each attribute value with its key and concatenated the attributes. For example, examples in COPA task have the following attributes: \textit{premise}, \textit{choice1}, \textit{choice2}, and \textit{question}. The concatenated input string is of the format: \\
\textit{"Premise: This is the example's premise. First choice: this is the value of choice1. Second choice: this is the value of choice2. What is the \{cause, effect\}?"} \\
Here the cause and effect are the two possible values of the attribute \textit{question}.

Examples in the WSC task contain two specifically marked texts within the input text. One span is a noun and the other a pronoun or a verb with the pronoun information implicitly included. Following the original T5 example\footnote{\href{https://github.com/google-research/text-to-text-transfer-transformer/blob/main/t5/data/preprocessors.py}{https://github.com/google-research/text-to-text-transfer-transformer/blob/main/t5/data/preprocessors.py}}, we indicate the first span by surrounding it with an asterisk on each side, and the second span by surrounding it with a hash symbol on each side.

\subsubsection{Generative tasks}
We tested SloT5 models on three generative tasks: lemmatization, summarization (two datasets), and text simplification.

For lemmatization (Lem) we used a part of the Slovene ssj500k dataset  \citep{ssj500k} included in the universal dependencies dataset. The model received an individual sentence on the input and was trained to generate the same sentence with every word lemmatized. Punctuation marks were included in the training and test sets, but we ignored them during the scoring.

For summarization, we used two news datasets: AutoSentiNews (ASN) \citep{buvcar2018annotated} and Slovene Press Agency (STA) news \citep{vzagar2022cross}, extracted from the Gigafida corpus.
We fine-tuned and evaluated the T5 models on each dataset separately, treating each as a separate task. During fine-tuning, the input to a T5 model was an article text, and the output was its summary. 

Text simplification task aims to simplify the input text to increase its readability. Common strategies include splitting long, complex sentences into multiple shorter, simple sentences, and replacing complex words with simpler, commonly used words. We utillized the Slovene text simplification dataset SloTS \citep{slots}, which contains sentence-aligned complex texts (original) and their simplified versions. The dataset contains entries, where a single complex sentence is repeated several times, each time paired with a different simple sentence simplifying a part of the complex sentence. We merged all such entries into a single instance, containing the complex sentence and concatenated simplified sentences. For example, three entries $\left[(c_1, s_1), (c_1, s_2), (c_1, s_3)\right]$ were merged into one entry $\left[(c_1, s_1 s_2 s_3)\right]$. 

\subsection{Fine-tuning T5 and compared models}
\label{sec:finetuning}
We fine-tuned all compared T5 models (Slovene and multilingual) end-to-end on each task separately, using the HuggingFace transformers library \footnote{\href{https://huggingface.co/}{https://huggingface.co/}}. We used the AdamW optimizer with the batch size of 64. We saved the fine-tuned model after each epoch and selected the one that performed best on the validation set, using the ROUGE-L metric \citep{lin-2004-rouge}. We used the greedy search decoding method for the output generation and limited the maximum number of tokens in the output. We chose the maximum output length based on the target text length, shorter for the classification tasks and longer for the generative tasks. The maximum output lengths and the number of fine-tuning epochs for each task are presented in Table~\ref{tab:evalparams}. The complete code of our experiments is publicly available\footnote{\href{https://github.com/MatejUlcar/SloT5-tools}{https://github.com/MatejUlcar/SloT5-tools}}.

\begin{table}[htb]
\centering
    \caption{Evaluation parameters and performance metrics for seq2seq and BERT models for each of the datasets.}
    \label{tab:evalparams}
    \begin{tabular}{lrrrll}
     & \multicolumn{2}{c}{Epochs} & Output len. & & \\
    Task & seq2seq & BERT & (seq2seq) & Metric & Dataset\\
    \midrule
    BoolQ & 10 & 100 & 4 & accuracy & \citep{slovenesuperglue} \\
    CB & 15 & 100 & 6 & $F_1$ & \citep{slovenesuperglue} \\
    COPA & 15 & 100 & 6 & accuracy & \citep{slovenesuperglue} \\
    RTE & 15 & 100 & 6 & accuracy & \citep{slovenesuperglue} \\
    WSC & 20 & 100 & 6 & accuracy & \citep{slovenesuperglue} \\
    NER & 20 & 3 & 64 & $F_1$ & \citep{ssj500k} \\
    SA & 10 & 10 & 5 & $F_1$ & \citep{twittersentiment} \\
    Lem & 15 & - & 512 & word/sent. acc. & \citep{ssj500k} \\
    STA & 5 & - & 512 & ROUGE L & \citep{gigafida} \\
    ASN & 5 & - & 512 & ROUGE L & \citep{autosentinews} \\
    SloTS & 64 & - & 256 & ROUGE L & \citep{slots} \\
    \bottomrule
    \end{tabular}
\end{table}

We compared SloT5 models with multilingual mT5 models \citep{xue2021mt5}, multilingual mBART-50 model \citep{tang-etal-2021-multilingual}, and with four encoder BERT-like models (described below). We fine-tuned the mT5 and mBART-50 models in the exact same manner as the SloT5 models on all ten tasks. 
BERT-like models were fine-tuned on seven classification tasks, but not on the generative tasks (Lem, ASN, STA, SloTS), as they cannot generate text. \cite{zagar2022slovene} evaluated BERT-like models on the Slovene SuperGLUE benchmark. The BERT models were fine-tuned on each task individually for 100 epochs using the Jiant tool \citep{phang2020jiant} with the initial learning rate of $10^{-5}$. \cite{ulvcar2021sloberta} evaluated BERT models supporting Slovene on NER and SA tasks. They added a softmax classification layer on top of the BERT models and fine-tuned them for 3 epochs on the NER task with a batch size of 8, and for 10 epochs on the SA task with a batch size of 32.

\subsection{Quantitative results}
\label{sec:results}

We compared the results of our three monolingual SloT5 models (described in \Cref{sec:SloT5}) on the eleven tasks (described in \Cref{sec:tasks}) with two multilingual T5 models of comparable sizes: mT5-small and mT5-large \citep{xue2021mt5}, and with a multilingual BART model (mBART-50-large) \citep{tang-etal-2021-multilingual}. Due to their larger vocabulary sizes, mT5 models have many more parameters than comparable SloT5 models (300M vs. 60M for small and 1.2B vs. 750M for large). However, the transformer layers are identical in their number and size for both small models and for both large models. mBART-50-large model has 611M parameters, 12 encoder and 12 decoder layers, thus it lies somewhere between small and large T5 model concerning size.

For the classification tasks, we also compared the results with four encoder BERT-like models: multilingual BERT model (mBERT), multilingual XLM-RoBERTa model (XLM-R), trilingual CroSloEngual model (CSE, Croatian-Slovene-English) \citep{ulcar2020xlbert}, and monolingual Slovene RoBERTa model (SloBERTa).


\subsubsection{Classification tasks}

The evaluation results on classification tasks are presented in Table~\ref{tab:results1}. Some T5 models score 0 on certain SuperGLUE tasks. The reason is that the tasks were reformatted as generative tasks, and we check whether the generated text is equal to any of the class labels (in case of 0, it was not). We perform only minor post-process filtering of the generated texts, such as removing \texttt{<extra\_id\_0>} tokens added by the T5 models.

On SuperGLUE tasks, all seq2seq models perform poorly. While they do outperform BERT-like models on BoolQ and RTE tasks, they barely beat the majority classifier on both tasks. The exception is the mBART-50-large model, which lags behind the majority classifier on BoolQ, and mT5-small, which performs worse than majority classifier on RTE. On RTE, T5-sl-large\textsubscript{5} and T5-sl-small\textsubscript{5} perform the best of all evaluated models on this task.

On NER, the multilingual mT5-small model performs poorly, while mT5-large is the best T5 model. All the T5 models lag behind the BERT-like models on the NER task. The dataset used for the SA task has a low inter-annotator agreement, limiting the overall performance. The best performing T5 models on the SA task, mT5-large and T5-sl-large\textsubscript{5} perform on par and are only slightly worse than the best model on this task, SloBERTa. The small Slovene T5 models perform on par with multilingual BERT models on the SA task.

\begin{table}[htb]
\centering
    \caption{Results of the compared T5 and BERT models on classification tasks. The metric for each task is shown in Table~\ref{tab:evalparams}. The results of the best performing model for each task are in bold, the results of the best performing seq2seq model are underlined.}
   \label{tab:results1}
   \begin{tabular}{lccccccc}
    Model & BoolQ & CB & COPA & RTE & WSC & NER & SA \\
    \midrule
    Maj. class. & 63.3 & 21.7 & 50.0 & 58.6 & 65.8 & 0.0 & 20.3 \\
    T5-sl-small\textsubscript{1} & 66.6 & 0.0 & 47.6 & 58.6 & 47.9 & 48.1 & 57.4 \\
    T5-sl-small\textsubscript{5} & 66.6 & 0.0 & 50.0 & \textbf{\underline{65.5}} & \underline{65.1} & 66.0 & 60.4 \\
    T5-sl-large\textsubscript{1} & \textbf{\underline{70.0}} & 48.8 & \underline{51.6} & 58.6 & 61.6 & 53.2 & 59.2 \\
    T5-sl-large\textsubscript{3} & 60.0 & 50.9 & 48.8 & 62.1 & 64.3 & 70.8 & 61.0  \\
    T5-sl-large\textsubscript{5} & 63.3 & \underline{62.2} & 50.6 & \textbf{\underline{65.5}} & 59.6 & 66.9 & 61.8 \\
    mT5-small & \textbf{\underline{70.0}} & 0.0 & 0.0 & 55.2 & 0.0 & 2.7 & 56.6  \\
    mT5-large & \textbf{\underline{70.0}} & 0.0 & 0.0 & 58.6 & 45.2 & \underline{79.0} & \underline{61.9} \\
    mBART-50-large & 56.7 & 48.6 & 50.0 & 62.1 & \underline{65.1} & 74.7 & 54.7 \\
    mBERT & 63.3 & 65.1 & 54.4 & 57.9 & 61.6 & 88.5 & 57.6 \\
    XLM-R & 63.3 & 62.0 & 51.4 & 42.8 & 65.8 & 91.2 & 60.4 \\
    CSE BERT & 63.3 & 59.8 & 55.0 & 53.8 & 56.2 & 92.8 & 61.0 \\
    SloBERTa & 63.3 & \textbf{68.6} & \textbf{58.2} & 49.6 & \textbf{73.3} & \textbf{93.3} & \textbf{62.3} \\
    \bottomrule
  \end{tabular}
\end{table}

\subsubsection{Generative tasks}
The results on the generative tasks (lemmatization, two summarization tasks, and text simplification) are shown in Table~\ref{tab:results2}.
\begin{table}[htb]
\centering
    \caption{Results of the compared seq2seq models on generative tasks: lemmatization (Lem), two summarization tasks (STA and ASN), and text simplification (TS). The metric for each task is shown in Table~\ref{tab:evalparams}. The results of the best performing model for each task are in bold.}
   \label{tab:results2}
   \begin{tabular}{lcccc}
    Model & Lem & STA & ASN & TS \\
    \midrule
    T5-sl-small\textsubscript{1} & 90.3/37.1 & 20.6 & 20.0 & 29.1 \\
    T5-sl-small\textsubscript{5} & 95.6/62.2 & 22.1 & 22.3 & 33.2 \\
    T5-sl-large\textsubscript{1} & 95.5/58.5 & 21.5 & 21.9 & 28.3 \\
    T5-sl-large\textsubscript{3} & 97.2/72.2 & 25.0 & 22.7 & 30.4 \\
    T5-sl-large\textsubscript{5} & 97.0/73.6 & 25.3 & 22.5 & 30.2 \\
    mT5-small & 90.0/25.5 & 17.9 & 19.2 & 26.3 \\
    mT5-large & \textbf{98.3/75.6} & 24.5 & 23.2 & \textbf{35.3} \\
    mBART-50-large & 97.8/73.2 & \textbf{27.6} & \textbf{24.1} & 33.7 \\
    \bottomrule
  \end{tabular}
\end{table}

While mBART-50-large does not perform very well on the classification tasks, it is the best performing model on two out of four generative tasks, and the second best model on the other two tasks.
If we compare only T5 models, large models consistently outperform small models, when trained on the same amount of data. The difference in performance is especially notable for the mT5 models, while it is not as big for the SloT5 models. In general, the difference in performance between T5 models is the same as observed on NER and SA tasks: mT5-large performs the best (excluding mBART), followed by T5-sl-large, T5-sl-small, while mT5-small performs the worst. 

The difference in training time has a large impact for the T5-sl-small model, as with more training the performance improves significantly on most tasks, especially the generative tasks. T5-sl-small\textsubscript{5} outperforms T5-sl-large\textsubscript{1} on most tasks, the exceptions are BoolQ, CB, and COPA. While longer training does improve the performance of T5-sl-large model, the difference is modest and most noticeable between one and three epochs. Surprisingly, on the tested datasets, more training of large models does not always help and it is unclear whether T5-sl-large\textsubscript{3} or T5-sl-large\textsubscript{5} is the best performing SloT5 model; the results depend on the task.

\subsection{Qualitative analysis}
\label{sec:analysis}
In general, quantitative results are less informative for generative tasks compared to classification tasks. The main reason is that  the evaluation metrics such as ROUGE-L score are not strongly correlated with human judgements. Below, we provide qualitative analysis of the text simplification and summarization results, while for lemmatization we did not notice any significant patterns.

\subsubsection{Text simplification}
We qualitatively analyzed the four best performing models on the text simplification task, according to the ROUGE-L metric. We show selected examples from the test set of SloTS dataset in Table~\ref{tab:slots}. We selected examples where at least some of the models generate a reasonable simplification, and where noticeable and interesting difference between the models can be observed. The examples along with the models' outputs were translated into English, trying to mimic the original as best as possible, including the mistakes (where present). The original Slovene examples are shown in \Cref{tab:slots-sl} in Appendix.

One of the difficulties of the SloTS text simplification dataset is that many complex texts are archaic and sometimes poetic. The simplified text in the dataset tends to be written in more contemporary standard language. Such examples are very difficult for the seq2seq models to simplify and they mostly generate extractive summaries, leaving out adjectives and subordinate clauses. This can best be seen in the first and the last two examples in Table~\ref{tab:slots}. Another large issue in this task is the hallucination, as all the models frequently invent information not present in the original sentence. This is most commonly the case with mBART-50-large model, which is the second best performing model, according to the ROUGE-L score. T5-sl-small\textsubscript{5} is the most robust model on this task. Compared to other models, it the most consistently produces coherent and truthful simplifications, though it still often invents new information. On the other hand, it most frequently generates shorter outputs, leaving out information in subordinate clauses. On the examples, where all the models fail, T5-sl-small\textsubscript{5} tends to perform the worst. mT5-large achieves the best ROUGE-L score. When examining its outputs, however, it seems that it either works very well (the first and the last example in Table~\ref{tab:slots}) or completely fails to produce meaningful or even grammatically correct sentences (the third and the fourth example in Table~\ref{tab:slots}).

When dealing with a relatively simple example with neutral language (example 2 in Table~\ref{tab:slots}), all models perform very well. However, on a more complex and longer example in the same domain (not shown to save space), none of the models produce a meaningful simplification.

\begin{table}[!htb]
    \caption{Examples of generated simplified sentences on the text simplification task for the best performing models, translated into English. Grammatical and other mistakes in the model outputs were intentionally kept.}
   \label{tab:slots}
   \begin{tabular}{lp{0.7\textwidth}}
    \toprule
    \multicolumn{2}{p{0.9\textwidth}}{When they noticed Matic, who was ringing the bells with all his might, they suddenly stopped and looked at him wide-eyed then asked him: "Matic, where is the fire?"} \\
    \midrule
    golden & They saw that Enaka Palica was ringing. People asked: where is the fire? \\
    mT5-large & When people were watching Matic, they got very angry. They were asking him: Matic, where is the fire? \\
    mBART-50-large & The judges decided, where it burns. \\
    T5-sl-small\textsubscript{5} & They asked Matic: Where's the fire? \\
    T5-sl-large\textsubscript{3} & When they noticed Matic, they looked at him wide-eyed. \\
    \midrule
    \multicolumn{2}{p{\textwidth}}{With the access ramp, the miniature train is accessible also for people using wheelchairs.} \\
    \midrule
    golden & People with wheelchairs can also ride with Urban. \\
    mT5-large & The miniature train is also accessible to people using wheelchairs. \\
    mBART-50-large & People using wheelchairs can get also an access ramp. \\
    T5-sl-small\textsubscript{5} & The miniature train is accessible also to disabled people. \\
    T5-sl-large\textsubscript{3} & The miniature train is accessible also to people, who have a wheelchair. \\
    \midrule
    \multicolumn{2}{p{\textwidth}}{And because the branches were swinging, so was the Moon above them, which was hanging from the sky. It was really bloody, and no longer fully round, and it was hanging low, like a lonely paper lantern, beaten, yet still burning above an empty fairground.} \\
    \midrule
    golden & The Moon was coloured red. \\
    mT5-large & He was bloody. \\
    mBART-50-large & The fair was over. Now, the father has returned to the stand. \\
    T5-sl-small\textsubscript{5} & At the fair was also the Moon, which was very bloody. \\
    T5-sl-large\textsubscript{3} & It has never haunted again. \\
    \midrule
    \multicolumn{2}{p{\textwidth}}{Above the pile of the old wall, covered with blackberries, clematis, and other shrubs, only the former castle tower was standing tall, proudly braving the hundred-year storms and gales.} \\
    \midrule
    golden & An old tower also remained. \\
    mT5-large & On tower was many cstling plants. \\
    mBART-50-large & In the older part was also the castle tower. \\
    T5-sl-small\textsubscript{5} & There was a large tower on the courtyard. \\
    T5-sl-large\textsubscript{3} & There used to be only a tower in the town. \\
    \midrule
    \multicolumn{2}{p{\textwidth}}{After that I was storming around Germany and taking part in battles, which were truly beautiful.} \\
    \midrule
    golden & I liked to battle. \\
    mT5-large & Along the way I met German of soldiers. \\
    mBART-50-large & After that I was storming around Germany and taking part in battles. \\
    T5-sl-small\textsubscript{5} & I took part in battles in Germany. \\
    T5-sl-large\textsubscript{3} & I strolled a lot around Germany. \\
    \midrule
    \multicolumn{2}{p{\textwidth}}{The baby was at that time already two years old and his mother was there with him on the field above the road, where she was hoeing.} \\
    \midrule
    golden & At that time the boy was 2 years old. Terba was digging on the field. The boy was along with her. \\
    mT5-large & Mother was with him at home. She was digging and hoeing. \\
    mBART-50-large & Mother was then already two years old. \\
    T5-sl-small\textsubscript{5} & Baby was very old. \\
    T5-sl-large\textsubscript{3} & There was a large baby on the field. \\
    \bottomrule
  \end{tabular}
\end{table}

\subsubsection{Summarization}
Below we summarize the qualitative findings concerning the results of different models on the ASR summarization dataset. Due to their length, we defer a few illustrative examples to Appendix (see the original Slovene examples in Table \ref{tab:asn-examples-sl} and their translations into English in Table  \ref{tab:asn-examples-en}).

In our summarization tasks, the ROUGE-L scores (measuring the longest n-gram overlap between the golden and generated summary) do no accurately represent the quality of the generated summaries. The generated summaries may be correct, concise and easy to read, yet the scores are low, because they focused on a different aspect of the news article than the provided golden summary. Commonly, the provided golden summary rounds or approximates the numbers and heavily paraphrases the text. The summaries generated by seq2seq models do not round the numbers and frequently copy whole sentences from the original news article. When they do paraphrase the text, they usually do it differently than the golden summaries.

A smaller T5-sl-small\textsubscript{5} model commonly generates summary only from the beginning or at most the first half of the article. It is also more prone to copying whole sentences from the input text. The biggest issue of T5-sl-small\textsubscript{5} is mixing up factual indicators, e.g., increase vs. decrease, most vs. least. It also tends to invent named entities, especially locations, putting many events in Ljubljana, the capital and the largest city of Slovenia, especially when no location is indicated in the original article (see the first example in \Cref{tab:asn-examples-en} in Appendix). Occasionally, the model is unable to form a coherent summary.

The mBART-50-large model has similar issues as T5-sl-small\textsubscript{5}, but on a smaller scale. It tends not to mix factual indicators. It is the most robust model in the sense that it most frequently produces a summary that conveys the crucial information in the article. It does so by frequently copying one or two input sentences it identifies as the most important to the output and only slightly modifies them. However, it does have issues with named entities, mostly leaving them out of the generated summaries. Thus, we often can not tell who did what or where, just that something was done (see the second and third examples in \Cref{tab:asn-examples-en} in Appendix). Similarly to the simplification task, mBART-50-large has the largest tendency to hallucinate (in this case invent wrong named entities) among the analyzed models (see the first example in \Cref{tab:asn-examples-en}). It is also the model with the largest number of grammatical mistakes. When omitting subordinate clauses, adjectives and verbs, or changing the verb, the noun declensions and/or verb conjugations should also be changed to fit the new sentence, but the model leaves them in the same form as in the input text.

T5-sl-large\textsubscript{3} and mT5-large both tend to generate good summaries. While we can observe many differences between the generated summaries of the two models, we can not point out any significant qualitative differences. The differences are mainly stylistic or due to chance. T5-sl-large\textsubscript{3} paraphrases the text more often (see the third example in \Cref{tab:asn-examples-en}). mT5-large, on the other hand, has more closely adapted to the summary format of the golden summaries.

\section{Discussion and conclusions}
\label{sec:discussion}

We presented three new T5 seq2seq models for Slovene. Our comparison of monolingual and multilingual T5 and BERT-based models, applicable to Slovene, shows that in general, for classification tasks, BERT models are preferable, while for text generation tasks, T5 models show reasonable performance. The specific findings are elaborated on below. While the results are obtained on Slovene, we believe that they may generalize to other less-resourced languages, where such models will be built. We make the training and evaluation code, as well as the trained models publicly available. The code can be found at \href{https://github.com/MatejUlcar/SloT5-tools}{https://github.com/MatejUlcar/SloT5-tools}.
The released models can be found at \href{https://www.huggingface.co/cjvt}{https://www.huggingface.co/cjvt}.

Both small Slovene T5 models outperform the multilingual small T5 model. However, the large multilingual model outperforms the large Slovene T5 model. Since T5-sl-small\textsubscript{1} and T5-sl-large\textsubscript{1} were trained for an equal amount of steps, we assume that the larger model is under-trained. \cite{1epoch} and \cite{hoffmann2022training} have recently presented evidence that the amount of training needs to scale with the size of the model. However, there is no consensus on the optimal amount of training required for a given model architecture. \cite{1epoch} suggests that given a fixed number of FLOPS (floating point operations per second) the optimal ratio between the number of training tokens and the number of model parameters is around 5. \cite{hoffmann2022training} on the other hand, report that ratio should be larger, around 20. For our T5-sl-large\textsubscript{1} model, this ratio is $5.5$, for T5-sl-large\textsubscript{3} $20$, for T5-sl-large\textsubscript{5} $33$, for T5-sl-small\textsubscript{1} $68$ and for T5-sl-small\textsubscript{5} $414$.

T5-sl-small\textsubscript{5} and T5-sl-large\textsubscript{1} were trained using roughly equal amount of computing power. Since T5-sl-small\textsubscript{5} outperforms T5-sl-large\textsubscript{1} on most tasks, we conclude that the optimal ratio between the number of training tokens and model parameters must be higher than 5 for Slovenian T5 models.

We observe that T5-sl-small strongly outperforms multilingual mT5-small. On the other hand, mT5-large performs better than T5-sl-large. Furthermore, while T5-sl-large\textsubscript{1} is clearly worse than T5-sl-large\textsubscript{3} and T5-sl-large\textsubscript{5}, the difference between the latter two is negligible. We hypothesize that the reason for better performance of the small SloT5 model, in comparison with the mT5, is that the small models have too few parameters to successfully encode (and decode) the information in multiple languages, so a monolingual model prevails. Our hypothesis for the worse performance of T5-sl-large, compared to mT5-large and mBART-50-large, is that there was not enough training data to successfully train a model of this size, especially since further training (for more epochs) does not seem to improve the performance. mT5-large was trained on a much larger training corpus, and even its Slovenian portion was almost twice as large as our corpus.

T5 and other seq2seq models can generate text, making them suitable for solving a wider variety of NLP tasks than encoder-only models, such as BERT. However, compared to BERT-like models, T5 models seem to be much more sensitive to unbalanced classes and smaller datasets. In addition to just classifying the input, the T5 models also have to learn how to form a coherent response. This is a simple task for a limited scope of available answers, such as most SuperGLUE classification tasks, but considerably different for the NER task, which we have formatted as the text retrieval task. Still, multilingual T5 models, especially mT5-small, have often failed in learning to generate even a sensible incorrect answer, i.e. predicting any class, even incorrect. Instead, they generate answers that are not identifiable with any class value.

Fine-tuning T5 models for more epochs on a specific task might solve the issue of generating nonsensical answers; however, we may over-fit the models. Furthermore, on models that did not have this problem, we have not observed a significant change in performance on the SuperGLUE tasks when training for more than 6-8 epochs.

Although the English T5 models \citep{raffel2020exploring} were pre-trained on multiple tasks, including the SuperGLUE tasks, the authors fine-tuned the pre-trained models for each task during the evaluation. Their results show that the largest T5 models achieve better results than the RoBERTa\textsubscript{LARGE} \citep{liu2019roberta} baseline. However, those models are of an order of magnitude larger than the baseline model. Comparing the performance of similarly sized models, the RoBERTa model outperforms T5 on all SuperGLUE tasks. \cite{xue2021mt5} reported much better performance of mT5 models compared to multilingual BERT-like models in the zero-shot cross-lingual setting. In a monolingual setting, only the largest (3B and 11B) mT5 models outperform mBERT on the NER task. On the other hand, on the question answering task, all mT5 models (except for the smallest mT5-small) outperform the mBERT score. This is in line with our findings, where we observe a slight improvement of T5 models over BERT-like models on the question answering BoolQ task but worse performance on other SuperGLUE tasks. 

In future work, we will try to obtain more Slovene data and retrain the large Slovene T5 model to  analyze the behaviour of the generative models with respect to the size of the training data. As text generation seems to be a stronger side of T5 models, we will expand the set of tackled tasks to paraphrasing and grammar correction tasks.

\section*{Funding}
The work was partially supported by the Slovenian Research Agency (ARRS) core research programme P6-0411 and projects J6-2581, J7-3159 and J1-2480, as well as the Ministry of Culture of Republic of Slovenia through project Development of Slovene in Digital Environment (RSDO).

\section*{Acknowledgments}
We acknowledge the efforts of SLING, Slovene national supercomputing grid for providing the necessary computational resources.

\bibliographystyle{model5-names}
\bibliography{bibs}

\begin{thebibliography}{45}
\expandafter\ifx\csname natexlab\endcsname\relax\def\natexlab#1{#1}\fi
\providecommand{\url}[1]{\texttt{#1}}
\providecommand{\href}[2]{#2}
\providecommand{\path}[1]{#1}
\providecommand{\DOIprefix}{doi:}
\providecommand{\ArXivprefix}{arXiv:}
\providecommand{\URLprefix}{URL: }
\providecommand{\Pubmedprefix}{pmid:}
\providecommand{\doi}[1]{\href{http://dx.doi.org/#1}{\path{#1}}}
\providecommand{\Pubmed}[1]{\href{pmid:#1}{\path{#1}}}
\providecommand{\bibinfo}[2]{#2}
\ifx\xfnm\relax \def\xfnm[#1]{\unskip,\space#1}\fi
\bibitem[{Bommasani et~al.(2021)Bommasani, Hudson, Adeli, Altman, Arora, von
  Arx, Bernstein, Bohg, Bosselut, Brunskill
  et~al.}]{bommasani2021opportunities}
\bibinfo{author}{Bommasani, R.}, \bibinfo{author}{Hudson, D.~A.},
  \bibinfo{author}{Adeli, E.}, \bibinfo{author}{Altman, R.},
  \bibinfo{author}{Arora, S.}, \bibinfo{author}{von Arx, S.},
  \bibinfo{author}{Bernstein, M.~S.}, \bibinfo{author}{Bohg, J.},
  \bibinfo{author}{Bosselut, A.}, \bibinfo{author}{Brunskill, E.} et~al.
  (\bibinfo{year}{2021}).
\newblock \bibinfo{title}{On the opportunities and risks of foundation models}.
\newblock {\it \bibinfo{journal}{ArXiv preprint 2108.07258}\/}, .
\bibitem[{Brown et~al.(2020)Brown, Mann, Ryder, Subbiah, Kaplan, Dhariwal,
  Neelakantan, Shyam, Sastry, Askell, Agarwal, Herbert-Voss, Krueger, Henighan,
  Child, Ramesh, Ziegler, Wu, Winter, Hesse, Chen, Sigler, Litwin, Gray, Chess,
  Clark, Berner, McCandlish, Radford, Sutskever \& Amodei}]{Brown2020GPT3}
\bibinfo{author}{Brown, T.}, \bibinfo{author}{Mann, B.},
  \bibinfo{author}{Ryder, N.}, \bibinfo{author}{Subbiah, M.},
  \bibinfo{author}{Kaplan, J.~D.}, \bibinfo{author}{Dhariwal, P.},
  \bibinfo{author}{Neelakantan, A.}, \bibinfo{author}{Shyam, P.},
  \bibinfo{author}{Sastry, G.}, \bibinfo{author}{Askell, A.},
  \bibinfo{author}{Agarwal, S.}, \bibinfo{author}{Herbert-Voss, A.},
  \bibinfo{author}{Krueger, G.}, \bibinfo{author}{Henighan, T.},
  \bibinfo{author}{Child, R.}, \bibinfo{author}{Ramesh, A.},
  \bibinfo{author}{Ziegler, D.}, \bibinfo{author}{Wu, J.},
  \bibinfo{author}{Winter, C.}, \bibinfo{author}{Hesse, C.},
  \bibinfo{author}{Chen, M.}, \bibinfo{author}{Sigler, E.},
  \bibinfo{author}{Litwin, M.}, \bibinfo{author}{Gray, S.},
  \bibinfo{author}{Chess, B.}, \bibinfo{author}{Clark, J.},
  \bibinfo{author}{Berner, C.}, \bibinfo{author}{McCandlish, S.},
  \bibinfo{author}{Radford, A.}, \bibinfo{author}{Sutskever, I.}, \&
  \bibinfo{author}{Amodei, D.} (\bibinfo{year}{2020}).
\newblock \bibinfo{title}{Language models are few-shot learners}.
\newblock In {\it \bibinfo{booktitle}{Advances in Neural Information Processing
  Systems}\/} (pp. \bibinfo{pages}{1877--1901}).
\newblock volume~\bibinfo{volume}{33}.
\bibitem[{Bu{\v c}ar(2017)}]{autosentinews}
\bibinfo{author}{Bu{\v c}ar, J.} (\bibinfo{year}{2017}).
\newblock \bibinfo{title}{Automatically sentiment annotated {Slovenian} news
  corpus {AutoSentiNews} 1.0}.
\newblock \URLprefix \url{http://hdl.handle.net/11356/1109}
  \bibinfo{note}{{Slovenian} language resource repository {CLARIN}.{SI}}.
\bibitem[{Bu{\v{c}}ar et~al.(2018)Bu{\v{c}}ar, {\v{Z}}nidar{\v{s}}i{\v{c}} \&
  Povh}]{buvcar2018annotated}
\bibinfo{author}{Bu{\v{c}}ar, J.},
  \bibinfo{author}{{\v{Z}}nidar{\v{s}}i{\v{c}}, M.}, \& \bibinfo{author}{Povh,
  J.} (\bibinfo{year}{2018}).
\newblock \bibinfo{title}{Annotated news corpora and a lexicon for sentiment
  analysis in slovene}.
\newblock {\it \bibinfo{journal}{Language Resources and Evaluation}\/},  {\it
  \bibinfo{volume}{52}\/}, \bibinfo{pages}{895--919}.
\bibitem[{Devlin et~al.(2019)Devlin, Chang, Lee \& Toutanova}]{Devlin2019}
\bibinfo{author}{Devlin, J.}, \bibinfo{author}{Chang, M.-W.},
  \bibinfo{author}{Lee, K.}, \& \bibinfo{author}{Toutanova, K.}
  (\bibinfo{year}{2019}).
\newblock \bibinfo{title}{{BERT}: Pre-training of deep bidirectional
  transformers for language understanding}.
\newblock In {\it \bibinfo{booktitle}{Proceedings of the 2019 Conference of the
  {N}orth {A}merican Chapter of the {A}ssociation for Computational
  Linguistics: Human Language Technologies, Volume 1 (Long and Short
  Papers)}\/} (pp. \bibinfo{pages}{4171--4186}).
\newblock \DOIprefix\doi{10.18653/v1/N19-1423}.
\bibitem[{Erjavec et~al.(2021)Erjavec, Fi{\v{s}}er \&
  Ljube{\v{s}}i{\'c}}]{erjavec2021kas}
\bibinfo{author}{Erjavec, T.}, \bibinfo{author}{Fi{\v{s}}er, D.}, \&
  \bibinfo{author}{Ljube{\v{s}}i{\'c}, N.} (\bibinfo{year}{2021}).
\newblock \bibinfo{title}{The {KAS} corpus of {Slovenian} academic writing}.
\newblock {\it \bibinfo{journal}{Language Resources and Evaluation}\/},  {\it
  \bibinfo{volume}{55}\/}, \bibinfo{pages}{551--583}.
\bibitem[{Erjavec et~al.(2017{\natexlab{a}})Erjavec, Ljube{\v s}i{\'c} \& Fi{\v
  s}er}]{11356/1138}
\bibinfo{author}{Erjavec, T.}, \bibinfo{author}{Ljube{\v s}i{\'c}, N.}, \&
  \bibinfo{author}{Fi{\v s}er, D.} (\bibinfo{year}{2017}{\natexlab{a}}).
\newblock \bibinfo{title}{Blog post and comment corpus {Janes-Blog} 1.0}.
\newblock \URLprefix \url{http://hdl.handle.net/11356/1138}
  \bibinfo{note}{{Slovenian} language resource repository {CLARIN}.{SI}}.
\bibitem[{Erjavec et~al.(2017{\natexlab{b}})Erjavec, Ljube{\v s}i{\'c} \& Fi{\v
  s}er}]{11356/1139}
\bibinfo{author}{Erjavec, T.}, \bibinfo{author}{Ljube{\v s}i{\'c}, N.}, \&
  \bibinfo{author}{Fi{\v s}er, D.} (\bibinfo{year}{2017}{\natexlab{b}}).
\newblock \bibinfo{title}{Forum corpus {Janes-Forum} 1.0}.
\newblock \URLprefix \url{http://hdl.handle.net/11356/1139}
  \bibinfo{note}{{Slovenian} language resource repository {CLARIN}.{SI}}.
\bibitem[{Erjavec et~al.(2017{\natexlab{c}})Erjavec, Ljube{\v s}i{\'c} \& Fi{\v
  s}er}]{11356/1140}
\bibinfo{author}{Erjavec, T.}, \bibinfo{author}{Ljube{\v s}i{\'c}, N.}, \&
  \bibinfo{author}{Fi{\v s}er, D.} (\bibinfo{year}{2017}{\natexlab{c}}).
\newblock \bibinfo{title}{News comment corpus {Janes-News} 1.0}.
\newblock \URLprefix \url{http://hdl.handle.net/11356/1140}
  \bibinfo{note}{{Slovenian} language resource repository {CLARIN}.{SI}}.
\bibitem[{Fi{\v{s}}er et~al.(2016)Fi{\v{s}}er, Erjavec \&
  Ljube{\v{s}}i{\'c}}]{fivser2016janes}
\bibinfo{author}{Fi{\v{s}}er, D.}, \bibinfo{author}{Erjavec, T.}, \&
  \bibinfo{author}{Ljube{\v{s}}i{\'c}, N.} (\bibinfo{year}{2016}).
\newblock \bibinfo{title}{{JANES v0. 4: Korpus} slovenskih spletnih
  uporabni{\v{s}}kih vsebin}.
\newblock {\it \bibinfo{journal}{Sloven{\v{s}}{\v{c}}ina 2.0: empirical,
  applied and interdisciplinary research}\/},  {\it \bibinfo{volume}{4}\/},
  \bibinfo{pages}{67--99}.
\bibitem[{Gorenc \& Robnik-{\v S}ikonja(2022)}]{slots}
\bibinfo{author}{Gorenc, S.}, \& \bibinfo{author}{Robnik-{\v S}ikonja, M.}
  (\bibinfo{year}{2022}).
\newblock \bibinfo{title}{Slovene text simplification dataset {SloTS}}.
\newblock \URLprefix \url{http://hdl.handle.net/11356/1682}
  \bibinfo{note}{{Slovenian} language resource repository {CLARIN}.{SI}}.
\bibitem[{Hoffmann et~al.(2022)Hoffmann, Borgeaud, Mensch, Buchatskaya, Cai,
  Rutherford, Casas, Hendricks, Welbl, Clark, Hennigan, Noland, Millican,
  Driessche, Damoc, Guy, Osindero, Simonyan, Elsen, Rae, Vinyals \&
  Sifre}]{hoffmann2022training}
\bibinfo{author}{Hoffmann, J.}, \bibinfo{author}{Borgeaud, S.},
  \bibinfo{author}{Mensch, A.}, \bibinfo{author}{Buchatskaya, E.},
  \bibinfo{author}{Cai, T.}, \bibinfo{author}{Rutherford, E.},
  \bibinfo{author}{Casas, D. d.~L.}, \bibinfo{author}{Hendricks, L.~A.},
  \bibinfo{author}{Welbl, J.}, \bibinfo{author}{Clark, A.},
  \bibinfo{author}{Hennigan, T.}, \bibinfo{author}{Noland, E.},
  \bibinfo{author}{Millican, K.}, \bibinfo{author}{Driessche, G. v.~d.},
  \bibinfo{author}{Damoc, B.}, \bibinfo{author}{Guy, A.},
  \bibinfo{author}{Osindero, S.}, \bibinfo{author}{Simonyan, K.},
  \bibinfo{author}{Elsen, E.}, \bibinfo{author}{Rae, J.~W.},
  \bibinfo{author}{Vinyals, O.}, \& \bibinfo{author}{Sifre, L.}
  (\bibinfo{year}{2022}).
\newblock \bibinfo{title}{Training compute-optimal large language models}.
\newblock {\it \bibinfo{journal}{ArXiv preprint 2203.15556}\/}, .
  \DOIprefix\doi{10.48550/ARXIV.2203.15556}.
\bibitem[{Komatsuzaki(2019)}]{1epoch}
\bibinfo{author}{Komatsuzaki, A.} (\bibinfo{year}{2019}).
\newblock \bibinfo{title}{One epoch is all you need}.
\newblock {\it \bibinfo{journal}{ArXiv preprint 1906.06669}\/}, .
  \DOIprefix\doi{10.48550/ARXIV.1906.06669}.
\bibitem[{Krek et~al.(2020)Krek, Arhar~Holdt, Erjavec, {\v{C}}ibej, Repar,
  Gantar, Ljube{\v{s}}i{\'c}, Kosem \& Dobrovoljc}]{krek-etal-2020-gigafida}
\bibinfo{author}{Krek, S.}, \bibinfo{author}{Arhar~Holdt, {\v{S}}.},
  \bibinfo{author}{Erjavec, T.}, \bibinfo{author}{{\v{C}}ibej, J.},
  \bibinfo{author}{Repar, A.}, \bibinfo{author}{Gantar, P.},
  \bibinfo{author}{Ljube{\v{s}}i{\'c}, N.}, \bibinfo{author}{Kosem, I.}, \&
  \bibinfo{author}{Dobrovoljc, K.} (\bibinfo{year}{2020}).
\newblock \bibinfo{title}{Gigafida 2.0: The reference corpus of written
  standard {S}lovene}.
\newblock In {\it \bibinfo{booktitle}{Proceedings of the 12th Language
  Resources and Evaluation Conference}\/} (pp. \bibinfo{pages}{3340--3345}).
\bibitem[{Krek et~al.(2019{\natexlab{a}})Krek, Dobrovoljc, Erjavec, Mo{\v z}e,
  Ledinek, Holz, Zupan, Gantar, Kuzman, {\v C}ibej, Arhar~Holdt, Kav{\v c}i{\v
  c}, {\v S}krjanec, Marko, Jezer{\v s}ek \& Zajc}]{ssj500k}
\bibinfo{author}{Krek, S.}, \bibinfo{author}{Dobrovoljc, K.},
  \bibinfo{author}{Erjavec, T.}, \bibinfo{author}{Mo{\v z}e, S.},
  \bibinfo{author}{Ledinek, N.}, \bibinfo{author}{Holz, N.},
  \bibinfo{author}{Zupan, K.}, \bibinfo{author}{Gantar, P.},
  \bibinfo{author}{Kuzman, T.}, \bibinfo{author}{{\v C}ibej, J.},
  \bibinfo{author}{Arhar~Holdt, {\v S}.}, \bibinfo{author}{Kav{\v c}i{\v c},
  T.}, \bibinfo{author}{{\v S}krjanec, I.}, \bibinfo{author}{Marko, D.},
  \bibinfo{author}{Jezer{\v s}ek, L.}, \& \bibinfo{author}{Zajc, A.}
  (\bibinfo{year}{2019}{\natexlab{a}}).
\newblock \bibinfo{title}{Training corpus ssj500k 2.2}.
\newblock \bibinfo{note}{{Slovenian} language resource repository
  {CLARIN}.{SI}}.
\bibitem[{Krek et~al.(2019{\natexlab{b}})Krek, Erjavec, Repar, {\v C}ibej,
  Arhar~Holdt, Gantar, Kosem, Robnik-{\v S}ikonja, Ljube{\v s}i{\'c},
  Dobrovoljc, Laskowski, Gr{\v c}ar, Holozan, {\v S}uster, Gorjanc, Stabej \&
  Logar}]{gigafida}
\bibinfo{author}{Krek, S.}, \bibinfo{author}{Erjavec, T.},
  \bibinfo{author}{Repar, A.}, \bibinfo{author}{{\v C}ibej, J.},
  \bibinfo{author}{Arhar~Holdt, {\v S}.}, \bibinfo{author}{Gantar, P.},
  \bibinfo{author}{Kosem, I.}, \bibinfo{author}{Robnik-{\v S}ikonja, M.},
  \bibinfo{author}{Ljube{\v s}i{\'c}, N.}, \bibinfo{author}{Dobrovoljc, K.},
  \bibinfo{author}{Laskowski, C.}, \bibinfo{author}{Gr{\v c}ar, M.},
  \bibinfo{author}{Holozan, P.}, \bibinfo{author}{{\v S}uster, S.},
  \bibinfo{author}{Gorjanc, V.}, \bibinfo{author}{Stabej, M.}, \&
  \bibinfo{author}{Logar, N.} (\bibinfo{year}{2019}{\natexlab{b}}).
\newblock \bibinfo{title}{Corpus of written standard {Slovene Gigafida} 2.0}.
\newblock \URLprefix \url{http://hdl.handle.net/11356/1320}
  \bibinfo{note}{{Slovenian} language resource repository {CLARIN}.{SI}}.
\bibitem[{Lewis et~al.(2020)Lewis, Liu, Goyal, Ghazvininejad, Mohamed, Levy,
  Stoyanov \& Zettlemoyer}]{lewis-etal-2020-bart}
\bibinfo{author}{Lewis, M.}, \bibinfo{author}{Liu, Y.}, \bibinfo{author}{Goyal,
  N.}, \bibinfo{author}{Ghazvininejad, M.}, \bibinfo{author}{Mohamed, A.},
  \bibinfo{author}{Levy, O.}, \bibinfo{author}{Stoyanov, V.}, \&
  \bibinfo{author}{Zettlemoyer, L.} (\bibinfo{year}{2020}).
\newblock \bibinfo{title}{{BART}: Denoising sequence-to-sequence pre-training
  for natural language generation, translation, and comprehension}.
\newblock In {\it \bibinfo{booktitle}{Proceedings of the 58th Annual Meeting of
  the Association for Computational Linguistics}\/} (pp.
  \bibinfo{pages}{7871--7880}).
\newblock \DOIprefix\doi{10.18653/v1/2020.acl-main.703}.
\bibitem[{Lin(2004)}]{lin-2004-rouge}
\bibinfo{author}{Lin, C.-Y.} (\bibinfo{year}{2004}).
\newblock \bibinfo{title}{{ROUGE}: A package for automatic evaluation of
  summaries}.
\newblock In {\it \bibinfo{booktitle}{Text Summarization Branches Out}\/} (pp.
  \bibinfo{pages}{74--81}).
\bibitem[{Liu et~al.(2020)Liu, Gu, Goyal, Li, Edunov, Ghazvininejad, Lewis \&
  Zettlemoyer}]{liu-etal-2020-multilingual}
\bibinfo{author}{Liu, Y.}, \bibinfo{author}{Gu, J.}, \bibinfo{author}{Goyal,
  N.}, \bibinfo{author}{Li, X.}, \bibinfo{author}{Edunov, S.},
  \bibinfo{author}{Ghazvininejad, M.}, \bibinfo{author}{Lewis, M.}, \&
  \bibinfo{author}{Zettlemoyer, L.} (\bibinfo{year}{2020}).
\newblock \bibinfo{title}{{Multilingual Denoising Pre-training for Neural
  Machine Translation}}.
\newblock {\it \bibinfo{journal}{Transactions of the Association for
  Computational Linguistics}\/},  {\it \bibinfo{volume}{8}\/},
  \bibinfo{pages}{726--742}. \DOIprefix\doi{10.1162/tacl_a_00343}.
\bibitem[{Liu et~al.(2019)Liu, Ott, Goyal, Du, Joshi, Chen, Levy, Lewis,
  Zettlemoyer \& Stoyanov}]{liu2019roberta}
\bibinfo{author}{Liu, Y.}, \bibinfo{author}{Ott, M.}, \bibinfo{author}{Goyal,
  N.}, \bibinfo{author}{Du, J.}, \bibinfo{author}{Joshi, M.},
  \bibinfo{author}{Chen, D.}, \bibinfo{author}{Levy, O.},
  \bibinfo{author}{Lewis, M.}, \bibinfo{author}{Zettlemoyer, L.}, \&
  \bibinfo{author}{Stoyanov, V.} (\bibinfo{year}{2019}).
\newblock \bibinfo{title}{{RoBERTa: A} robustly optimized {BERT} pretraining
  approach}.
\newblock {\it \bibinfo{journal}{ArXiv preprint 1907.11692}\/}, .
\bibitem[{Ljube{\v{s}}i{\'c} \& Erjavec(2011)}]{ljubevsic2011hrwac}
\bibinfo{author}{Ljube{\v{s}}i{\'c}, N.}, \& \bibinfo{author}{Erjavec, T.}
  (\bibinfo{year}{2011}).
\newblock \bibinfo{title}{{hrWaC} and {slWaC}: {Compiling} web corpora for
  {Croatian} and {Slovene}}.
\newblock In {\it \bibinfo{booktitle}{International Conference on Text, Speech
  and Dialogue}\/} (pp. \bibinfo{pages}{395--402}).
\bibitem[{Ljube{\v s}i{\'c} et~al.(2017)Ljube{\v s}i{\'c}, Erjavec \& Fi{\v
  s}er}]{11356/1137}
\bibinfo{author}{Ljube{\v s}i{\'c}, N.}, \bibinfo{author}{Erjavec, T.}, \&
  \bibinfo{author}{Fi{\v s}er, D.} (\bibinfo{year}{2017}).
\newblock \bibinfo{title}{Wikipedia talk corpus {Janes-Wiki} 1.0}.
\newblock \URLprefix \url{http://hdl.handle.net/11356/1137}
  \bibinfo{note}{{Slovenian} language resource repository {CLARIN}.{SI}}.
\bibitem[{Mozeti{\v{c}} et~al.(2016)Mozeti{\v{c}}, Gr{\v{c}}ar \&
  Smailovi{\'c}}]{mozetivc2016multilingual}
\bibinfo{author}{Mozeti{\v{c}}, I.}, \bibinfo{author}{Gr{\v{c}}ar, M.}, \&
  \bibinfo{author}{Smailovi{\'c}, J.} (\bibinfo{year}{2016}).
\newblock \bibinfo{title}{Multilingual {Twitter} sentiment classification: The
  role of human annotators}.
\newblock {\it \bibinfo{journal}{{PLOS ONE}}\/},  {\it \bibinfo{volume}{11}\/}.
\bibitem[{Mozeti{\v c} et~al.(2016)Mozeti{\v c}, Gr{\v c}ar \&
  Smailovi{\'c}}]{twittersentiment}
\bibinfo{author}{Mozeti{\v c}, I.}, \bibinfo{author}{Gr{\v c}ar, M.}, \&
  \bibinfo{author}{Smailovi{\'c}, J.} (\bibinfo{year}{2016}).
\newblock \bibinfo{title}{Twitter sentiment for 15 european languages}.
\newblock \URLprefix \url{http://hdl.handle.net/11356/1054}
  \bibinfo{note}{{Slovenian} language resource repository {CLARIN}.{SI}}.
\bibitem[{Nagoudi et~al.(2021)Nagoudi, Elmadany \&
  Abdul-Mageed}]{nagoudietal2021arat5}
\bibinfo{author}{Nagoudi, E. M.~B.}, \bibinfo{author}{Elmadany, A.}, \&
  \bibinfo{author}{Abdul-Mageed, M.} (\bibinfo{year}{2021}).
\newblock \bibinfo{title}{{AraT5: Text-to-Text Transformers for Arabic Language
  Generation}}, .
\newblock \URLprefix \url{https://arxiv.org/abs/2109.12068}.
  \DOIprefix\doi{10.48550/ARXIV.2109.12068}.
\bibitem[{Pan{\v{c}}ur \& Erjavec(2020)}]{pancur2020siparl}
\bibinfo{author}{Pan{\v{c}}ur, A.}, \& \bibinfo{author}{Erjavec, T.}
  (\bibinfo{year}{2020}).
\newblock \bibinfo{title}{The {siParl} corpus of {Slovene} parliamentary
  proceedings}.
\newblock In {\it \bibinfo{booktitle}{Proceedings of the Second ParlaCLARIN
  Workshop}\/} (pp. \bibinfo{pages}{28--34}).
\bibitem[{Pan{\v c}ur et~al.(2020)Pan{\v c}ur, Erjavec, Ojster{\v s}ek, {\v
  S}orn \& Blaj~Hribar}]{11356/1300}
\bibinfo{author}{Pan{\v c}ur, A.}, \bibinfo{author}{Erjavec, T.},
  \bibinfo{author}{Ojster{\v s}ek, M.}, \bibinfo{author}{{\v S}orn, M.}, \&
  \bibinfo{author}{Blaj~Hribar, N.} (\bibinfo{year}{2020}).
\newblock \bibinfo{title}{Slovenian parliamentary corpus (1990-2018) {siParl}
  2.0}.
\newblock \URLprefix \url{http://hdl.handle.net/11356/1300}
  \bibinfo{note}{{Slovenian} language resource repository {CLARIN}.{SI}}.
\bibitem[{Phang et~al.(2020)Phang, Yeres, Swanson, Liu, Tenney, Htut, Vania,
  Wang \& Bowman}]{phang2020jiant}
\bibinfo{author}{Phang, J.}, \bibinfo{author}{Yeres, P.},
  \bibinfo{author}{Swanson, J.}, \bibinfo{author}{Liu, H.},
  \bibinfo{author}{Tenney, I.~F.}, \bibinfo{author}{Htut, P.~M.},
  \bibinfo{author}{Vania, C.}, \bibinfo{author}{Wang, A.}, \&
  \bibinfo{author}{Bowman, S.~R.} (\bibinfo{year}{2020}).
\newblock \bibinfo{title}{\texttt{jiant} 2.0: A software toolkit for research
  on general-purpose text understanding models}.
\newblock \bibinfo{howpublished}{\url{http://jiant.info/}}.
\bibitem[{Pomik{\'a}lek(2011)}]{pomikalek2011removing}
\bibinfo{author}{Pomik{\'a}lek, J.} (\bibinfo{year}{2011}).
\newblock {\it \bibinfo{title}{Removing boilerplate and duplicate content from
  web corpora}\/}.
\newblock Ph.D. thesis Masaryk university, Brno, Czech Republic.
\bibitem[{Radford et~al.(2019)Radford, Wu, Child, Luan, Amodei \&
  Sutskever}]{Radford2019GPT2}
\bibinfo{author}{Radford, A.}, \bibinfo{author}{Wu, J.},
  \bibinfo{author}{Child, R.}, \bibinfo{author}{Luan, D.},
  \bibinfo{author}{Amodei, D.}, \& \bibinfo{author}{Sutskever, I.}
  (\bibinfo{year}{2019}).
\newblock {\it \bibinfo{title}{Language Models are Unsupervised Multitask
  Learners}\/}.
\newblock \bibinfo{type}{Technical Report} OpenAI blog. 2019 Feb 24.
\bibitem[{Raffel et~al.(2020)Raffel, Shazeer, Roberts, Lee, Narang, Matena,
  Zhou, Li \& Liu}]{raffel2020exploring}
\bibinfo{author}{Raffel, C.}, \bibinfo{author}{Shazeer, N.},
  \bibinfo{author}{Roberts, A.}, \bibinfo{author}{Lee, K.},
  \bibinfo{author}{Narang, S.}, \bibinfo{author}{Matena, M.},
  \bibinfo{author}{Zhou, Y.}, \bibinfo{author}{Li, W.}, \&
  \bibinfo{author}{Liu, P.~J.} (\bibinfo{year}{2020}).
\newblock \bibinfo{title}{Exploring the limits of transfer learning with a
  unified text-to-text transformer}.
\newblock {\it \bibinfo{journal}{Journal of Machine Learning Research}\/},
  {\it \bibinfo{volume}{21}\/}, \bibinfo{pages}{1--67}.
\bibitem[{Sarti \& Nissim(2022)}]{sarti-nissim2022it5}
\bibinfo{author}{Sarti, G.}, \& \bibinfo{author}{Nissim, M.}
  (\bibinfo{year}{2022}).
\newblock \bibinfo{title}{{IT5: Large-scale Text-to-text Pretraining for
  Italian Language Understanding and Generation}}, .
\newblock \DOIprefix\doi{10.48550/ARXIV.2203.03759}.
\bibitem[{Tang et~al.(2021)Tang, Tran, Li, Chen, Goyal, Chaudhary, Gu \&
  Fan}]{tang-etal-2021-multilingual}
\bibinfo{author}{Tang, Y.}, \bibinfo{author}{Tran, C.}, \bibinfo{author}{Li,
  X.}, \bibinfo{author}{Chen, P.-J.}, \bibinfo{author}{Goyal, N.},
  \bibinfo{author}{Chaudhary, V.}, \bibinfo{author}{Gu, J.}, \&
  \bibinfo{author}{Fan, A.} (\bibinfo{year}{2021}).
\newblock \bibinfo{title}{Multilingual translation from denoising
  pre-training}.
\newblock In {\it \bibinfo{booktitle}{Findings of the Association for
  Computational Linguistics: ACL-IJCNLP 2021}\/} (pp.
  \bibinfo{pages}{3450--3466}).
\newblock \DOIprefix\doi{10.18653/v1/2021.findings-acl.304}.
\bibitem[{Ul{\v{c}}ar \& Robnik-{\v{S}}ikonja(2021)}]{ulvcar2021sloberta}
\bibinfo{author}{Ul{\v{c}}ar, M.}, \& \bibinfo{author}{Robnik-{\v{S}}ikonja,
  M.} (\bibinfo{year}{2021}).
\newblock \bibinfo{title}{{SloBERTa}: Slovene monolingual large pretrained
  masked language model}.
\newblock In {\it \bibinfo{booktitle}{24th international multiconference
  Information Society 2021}\/}.
\newblock volume \bibinfo{volume}{C. Data Mining and Data Warehouses}.
\bibitem[{Ul{\v{c}}ar et~al.(2021)Ul{\v{c}}ar, {\v{Z}}agar, Armendariz, Repar,
  Pollak, Purver \& Robnik-{\v{S}}ikonja}]{ulcar2021evaluation}
\bibinfo{author}{Ul{\v{c}}ar, M.}, \bibinfo{author}{{\v{Z}}agar, A.},
  \bibinfo{author}{Armendariz, C.~S.}, \bibinfo{author}{Repar, A.},
  \bibinfo{author}{Pollak, S.}, \bibinfo{author}{Purver, M.}, \&
  \bibinfo{author}{Robnik-{\v{S}}ikonja, M.} (\bibinfo{year}{2021}).
\newblock \bibinfo{title}{Evaluation of contextual embeddings on less-resourced
  languages}.
\newblock {\it \bibinfo{journal}{Preprint arXiv:2107.10614}\/}, .
\bibitem[{Ulčar \& Robnik-Šikonja(2020)}]{ulcar2020xlbert}
\bibinfo{author}{Ulčar, M.}, \& \bibinfo{author}{Robnik-Šikonja, M.}
  (\bibinfo{year}{2020}).
\newblock \bibinfo{title}{{FinEst BERT and CroSloEngual BERT}: less is more in
  multilingual models.}
\newblock In {\it \bibinfo{booktitle}{Proceedings of {Text, Speech, and
  Dialogue, TSD} 2020}\/} (pp. \bibinfo{pages}{104--111}).
\bibitem[{Vaswani et~al.(2017)Vaswani, Shazeer, Parmar, Uszkoreit, Jones,
  Gomez, Kaiser \& Polosukhin}]{Vaswani2017}
\bibinfo{author}{Vaswani, A.}, \bibinfo{author}{Shazeer, N.},
  \bibinfo{author}{Parmar, N.}, \bibinfo{author}{Uszkoreit, J.},
  \bibinfo{author}{Jones, L.}, \bibinfo{author}{Gomez, A.~N.},
  \bibinfo{author}{Kaiser, {\L}.}, \& \bibinfo{author}{Polosukhin, I.}
  (\bibinfo{year}{2017}).
\newblock \bibinfo{title}{Attention is all you need}.
\newblock In {\it \bibinfo{booktitle}{Advances in neural information processing
  systems}\/} (pp. \bibinfo{pages}{5998--6008}).
\bibitem[{Wang et~al.(2019)Wang, Pruksachatkun, Nangia, Singh, Michael, Hill,
  Levy \& Bowman}]{wang2019superglue}
\bibinfo{author}{Wang, A.}, \bibinfo{author}{Pruksachatkun, Y.},
  \bibinfo{author}{Nangia, N.}, \bibinfo{author}{Singh, A.},
  \bibinfo{author}{Michael, J.}, \bibinfo{author}{Hill, F.},
  \bibinfo{author}{Levy, O.}, \& \bibinfo{author}{Bowman, S.~R.}
  (\bibinfo{year}{2019}).
\newblock \bibinfo{title}{{SuperGLUE: A} stickier benchmark for general-purpose
  language understanding systems}.
\newblock In {\it \bibinfo{booktitle}{Proceedings of the 33rd International
  Conference on Neural Information Processing Systems}\/}.
\bibitem[{Wang et~al.(2018)Wang, Singh, Michael, Hill, Levy \&
  Bowman}]{wang-etal-2018-glue}
\bibinfo{author}{Wang, A.}, \bibinfo{author}{Singh, A.},
  \bibinfo{author}{Michael, J.}, \bibinfo{author}{Hill, F.},
  \bibinfo{author}{Levy, O.}, \& \bibinfo{author}{Bowman, S.}
  (\bibinfo{year}{2018}).
\newblock \bibinfo{title}{{GLUE}: A multi-task benchmark and analysis platform
  for natural language understanding}.
\newblock In {\it \bibinfo{booktitle}{Proceedings of the 2018 {EMNLP} Workshop
  {B}lackbox{NLP}: Analyzing and Interpreting Neural Networks for {NLP}}\/}
  (pp. \bibinfo{pages}{353--355}).
\newblock \DOIprefix\doi{10.18653/v1/W18-5446}.
\bibitem[{Xue et~al.(2021)Xue, Constant, Roberts, Kale, Al-Rfou, Siddhant,
  Barua \& Raffel}]{xue2021mt5}
\bibinfo{author}{Xue, L.}, \bibinfo{author}{Constant, N.},
  \bibinfo{author}{Roberts, A.}, \bibinfo{author}{Kale, M.},
  \bibinfo{author}{Al-Rfou, R.}, \bibinfo{author}{Siddhant, A.},
  \bibinfo{author}{Barua, A.}, \& \bibinfo{author}{Raffel, C.}
  (\bibinfo{year}{2021}).
\newblock \bibinfo{title}{{mT5: A} massively multilingual pre-trained
  text-to-text transformer}.
\newblock In {\it \bibinfo{booktitle}{Proceedings of the 2021 Conference of the
  North American Chapter of the Association for Computational Linguistics:
  Human Language Technologies}\/} (pp. \bibinfo{pages}{483--498}).
\bibitem[{{\v Z}agar et~al.(2022){\v Z}agar, Kava{\v s}, Robnik-{\v S}ikonja,
  Erjavec, Fi{\v s}er, Ljube{\v s}i{\'c}, Ferme, Borovi{\v c}, Bo{\v s}kovi{\v
  c}, Ojster{\v s}ek \& Hrovat}]{11356/1448}
\bibinfo{author}{{\v Z}agar, A.}, \bibinfo{author}{Kava{\v s}, M.},
  \bibinfo{author}{Robnik-{\v S}ikonja, M.}, \bibinfo{author}{Erjavec, T.},
  \bibinfo{author}{Fi{\v s}er, D.}, \bibinfo{author}{Ljube{\v s}i{\'c}, N.},
  \bibinfo{author}{Ferme, M.}, \bibinfo{author}{Borovi{\v c}, M.},
  \bibinfo{author}{Bo{\v s}kovi{\v c}, B.}, \bibinfo{author}{Ojster{\v s}ek,
  M.}, \& \bibinfo{author}{Hrovat, G.} (\bibinfo{year}{2022}).
\newblock \bibinfo{title}{Corpus of academic {Slovene KAS} 2.0}.
\newblock \URLprefix \url{http://hdl.handle.net/11356/1448}
  \bibinfo{note}{{Slovenian} language resource repository {CLARIN}.{SI}}.
\bibitem[{{\v{Z}}agar \&
  Robnik-{\v{S}}ikonja(2022{\natexlab{a}})}]{vzagar2022cross}
\bibinfo{author}{{\v{Z}}agar, A.}, \& \bibinfo{author}{Robnik-{\v{S}}ikonja,
  M.} (\bibinfo{year}{2022}{\natexlab{a}}).
\newblock \bibinfo{title}{Cross-lingual transfer of abstractive summarizer to
  less-resource language}.
\newblock {\it \bibinfo{journal}{Journal of Intelligent Information
  Systems}\/},  {\it \bibinfo{volume}{58}\/}, \bibinfo{pages}{153--173}.
\bibitem[{{\v{Z}}agar \&
  Robnik-{\v{S}}ikonja(2022{\natexlab{b}})}]{zagar2022slovene}
\bibinfo{author}{{\v{Z}}agar, A.}, \& \bibinfo{author}{Robnik-{\v{S}}ikonja,
  M.} (\bibinfo{year}{2022}{\natexlab{b}}).
\newblock \bibinfo{title}{{Slovene {SuperGLUE} Benchmark: {Translation} and
  Evaluation}}.
\newblock In {\it \bibinfo{booktitle}{Proceedings of Language Resources and
  Evaluation, LREC}\/}.
\bibitem[{{\v Z}agar et~al.(2020){\v Z}agar, Robnik-{\v S}ikonja, Goli \&
  Arhar~Holdt}]{slovenesuperglue}
\bibinfo{author}{{\v Z}agar, A.}, \bibinfo{author}{Robnik-{\v S}ikonja, M.},
  \bibinfo{author}{Goli, T.}, \& \bibinfo{author}{Arhar~Holdt, {\v S}.}
  (\bibinfo{year}{2020}).
\newblock \bibinfo{title}{Slovene translation of {SuperGLUE}}.
\newblock \URLprefix \url{http://hdl.handle.net/11356/1380}
  \bibinfo{note}{{Slovenian} language resource repository {CLARIN}.{SI}}.
\bibitem[{Zhang et~al.(2021)Zhang, Zhang, Chen, Guo, Hua, Wang \&
  Zhou}]{zhangetal2021mengzi}
\bibinfo{author}{Zhang, Z.}, \bibinfo{author}{Zhang, H.},
  \bibinfo{author}{Chen, K.}, \bibinfo{author}{Guo, Y.}, \bibinfo{author}{Hua,
  J.}, \bibinfo{author}{Wang, Y.}, \& \bibinfo{author}{Zhou, M.}
  (\bibinfo{year}{2021}).
\newblock \bibinfo{title}{{Mengzi: Towards} lightweight yet ingenious
  pre-trained models for {Chinese}}, .
\newblock \DOIprefix\doi{10.48550/ARXIV.2110.06696}.

\end{thebibliography}

\appendix
\section{Additional examples}

\begin{table}[!htb]
    \caption{Original Slovene examples of generated simplified sentences on the text simplification task for the best performing models. The English translations are contained in Table \ref{tab:slots}.}
   \label{tab:slots-sl}
   \begin{tabular}{lp{0.8\textwidth}}
    \multicolumn{2}{p{1.0\textwidth}}{Ko so zagledali Matica, ki je zvonil na vse pretege, so najprej obstali in ga debelo pogledali, nato pa so butnili vanj: "Matic, kje gori?"} \\
    \midrule
    golden & Videli so, da zvoni Enaka Palica. Ljudje so vprašali: Kje gori? \\
    mT5-large & Ko so ljudje gledali Matica, so se zelo razjezili. Spraševali so ga: Matic, kje gori? \\
    mBART-50-large & Sodniki so se odločili, kje gori. \\
    T5-sl-small\textsubscript{5} & Matica so vprašali: Kje gori? \\
    T5-sl-large\textsubscript{3} & Ko so Matica zagledali, so ga debelo pogledali. \\
    \midrule
    \multicolumn{2}{p{\textwidth}}{S klančino je vlakec dostopen tudi osebam na invalidskih vozičkih.} \\
    \midrule
    golden & Z Urbanom se lahko peljejo ljudje na vozičkih. \\
    mT5-large & Vlakec je dostopen tudi osebam na invalidskih vozičkih. \\
    mBART-50-large & Ljudje na invalidskih vozičkih lahko dobijo tudi klančino. \\
    T5-sl-small\textsubscript{5} & Vlakec je dostopen tudi invalidom. \\
    T5-sl-large\textsubscript{3} & Vlakec je dostopen tudi osebam, ki imajo invalidski voziček. \\
    \midrule
    \multicolumn{2}{p{\textwidth}}{In ker so se zibale veje, se je nad njimi zibal tudi mesec, ki je visel izpod neba. Bil je res krvav in ne več popolnoma okrogel, in visel je nizko, kakor bi nad praznim sejmiščem svetil še edini, neugasli, a že obtolčeni lampijon.} \\
    \midrule
    golden & Luna je bila rdeče barve. \\
    mT5-large & Bil je krvav. \\
    mBART-50-large & Sejma je bilo konec. Zdaj se je oče vrnil na stojnico. \\
    T5-sl-small\textsubscript{5} & Na sejmu je bil tudi mesec, ki je bil zelo krvav. \\
    T5-sl-large\textsubscript{3} & Nikdar več ni strašilo. \\
    \midrule
    \multicolumn{2}{p{\textwidth}}{Iznad kupa starega zidovja, poraščenega z ostrogovno, srebotovino in drugim grmovjem, je molel edino še nekdanji grajski stolp kvišku, ki je ponosno kljuboval stoletnim viharjem in nevihtam.} \\
    \midrule
    golden & Ostal je tudi star stolp. \\
    mT5-large & Na stolp je bilo veliko grdnih rastlin. \\
    mBART-50-large & V starem delu je bil tudi grajski stolp. \\
    T5-sl-small\textsubscript{5} & Na dvorišču je bil velik stolp. \\
    T5-sl-large\textsubscript{3} & V mestu je bil včasih samo stolp. \\
    \midrule
    \multicolumn{2}{p{\textwidth}}{Nato pa sem vihral po Nemčiji okrog ter se udeleževal bitk, ki so bile res lepe.} \\
    \midrule
    golden & Rad sem se boril.\\
    mT5-large & Na poti sem srečal nemške vojakov. \\
    mBART-50-large & Nato sem vihral po Nemčiji in se udeleževal bitk. \\
    T5-sl-small\textsubscript{5} & Po Nemčiji sem se udeležil bitk. \\
    T5-sl-large\textsubscript{3} & Po Nemčiji sem se veliko sprehajal. \\
    \midrule
    \multicolumn{2}{p{\textwidth}}{Dete je bilo takrat že dve leti staro in mati je bila z njim na njivi nad cesto, kjer je okopavala.} \\
    \midrule
    golden & Takrat je bil fantek star 2 leti. Terba je kopala na njivi. Fantek je bil zraven. \\
    mT5-large & Mati je bila z njim doma. Kopala je in okopavala. \\
    mBART-50-large & Mati je bila takrat že dve leti staro. \\
    T5-sl-small\textsubscript{5} & Dete je bilo zelo staro. \\
    T5-sl-large\textsubscript{3} & Na njivi je bilo veliko dete. \\
    \bottomrule
  \end{tabular}
\end{table}

\begin{table}
\caption{Original Slovene examples of generated summaries on the ASN summarization task for the best performing models.}
\label{tab:asn-examples-sl}
\begin{tabular}{lp{0.8\textwidth}}
\multicolumn{2}{p{\textwidth}}{       Kandidati bodo o izboru obveščeni najpozneje do 3. aprila . Sklad je kot glavno željo prejel 60 vlog za stanovanja v Novem mestu , 53 za Postojno , 36 zaMaribor , 43 za Kranj in 15 vlog za stanovanja na Jesenicah . Povpraševanje je bilo najbolj izrazitoza lokacijah v Postojni in Novem mestu , kjer se sklad pojavlja prvič in je bil odziv zatopričakovan . Sklad bo prejete vloge podrobno preučil , izbor kupcev pa bo potekal 27. marca v navzočnostinotarja in komisije skladno z določbami razpisa . Stanovanja , ki bodo ostala neprodana v sklopu izbora , bo sklad objavil na svoji spletnistrani www.ssrs.si 3. aprila in jih sproti prodajal po pogojih razpisa od 7. aprila dalje . Kotkupci stanovanj v nadaljnji prodaji se lahko prijavijo vsi zainteresirani , ne le kupci iz izrecnonavedenih prednostnih kategorij .}\\
\midrule
golden &  Stanovanjski sklad RS je v roku za oddajo vlog za nakup 364 tržnih stanovanj sklada v Novem mestu , Postojni , Mariboru , Kranju in na Jesenicah , ki se je iztekel v ponedeljek , prejel 207 vlog .\\
mT5-large &       Sklad za stanovanja v Sloveniji je objavil razpis za prodajo stanovanj v nadaljnji prodaji.\\
mBART-50-large &        Sklad za financiranje razgradnje Jedrske elektrarne Krško ( JEK ) je objavil razpis za prodajo stanovanj v nadaljnji prodaji.\\
T5-sl-small\textsubscript{5} &  Ljubljana - Sklad za stanovanjsko gradnjo ( Sklad ) je danes objavil razpis za nakup stanovanj v Ljubljani, ki ga je sklad objavil v začetku marca.\\
T5-sl-large\textsubscript{3} &  Stanovanjski sklad RS je objavil razpis za prodajo stanovanj v nadaljnji prodaji stanovanj.\\
\midrule
\multicolumn{2}{p{\textwidth}}{Ivan Zidar iz SCT , prva dama Vegrada Hilda Tovšak in prvak Primorja Dušan Črnigoj še vedno lahko pristanejo v priporu . Prva dva je preiskovalna sodnica včeraj sicer spustila na prostost in zahtevala , da glede odreditve pripora odloči senat treh sodnikov Okrožnega sodišča v Ljubljani . Črnigoja so izpustili , ker naj bi se skliceval na imuniteto . Tožilstvo je danes v Državni svet že poslalo zahtevo , da se osumljencu imuniteto odvzame . Zidar in Tovšakova naj bi bila preveč radodarna Kazenska ovadba Zidarju , Tovšakovi in še trem drugim očita storitev kaznivega dejanja nedovoljenega dajanja daril po I. odstavku 248. člena Kazenskega zakonika , ki pravi : " Kdor osebi , ki opravlja gospodarsko dejavnost , obljubi , ponudi ali da nedovoljeno nagrado , darilo ali kakšno drugo korist zanjo ali za koga drugega zato , da bi sebi ali komu drugemu pridobil kakšno neupravičeno ugodnost pri sklenitvi posla ali storitvi iz prvega odstavka 247. člena tega zakonika ( nedovoljeno sprejemanje daril ) , se kaznuje z zaporom od šestih mesecev do petih let . V priporu pridržali le Tomaža Žiberta Kot smo že poročali je na današnji tiskovni konferenci vodja skupine tožilcev za pregon organiziranega kriminala pri Vrhovnem tožilstvu Harij Furlan povedal , da so pridržali sedem oseb , katerim očitajo dejanja dajanja in prejemanja daril . Preiskovalna sodnica je pripor odredila le članu razpisne komisije javnega podjetja Kontrola zračnega promet Tomažu Žibertu .}\\
\midrule
golden & Ivan Zidar in Hilda Tovšak , ki sta osumljena nedovoljenega dajanja daril , se lahko še vedno znajdeta v priporu , če tako odloči senat treh sodnikov . Črnigoju morajo prej odvzeti imuniteto\\
mT5-large &      Preiskovalna sodnica je včeraj spustila Ivana Zidarja, Hildo Tovšak in Dušana Črnigoja na prostost\\
mBART-50-large &     Tožilstvo je danes v Državni svet že poslalo zahtevo, da se osumljencu imuniteto odvzame\\
T5-sl-small\textsubscript{5} & Ivan Zidar, Hilda Tovšak in Dušan Črnigoj še vedno lahko pristanejo v priporu.\\
T5-sl-large\textsubscript{3} & Ivan Zidar, Hilda Tovšak in Dušan Črnigoj še vedno lahko pristanejo v priporu\\
\midrule
\multicolumn{2}{p{\textwidth}}{ Norveška policija preiskuje navedbe Andersa Behringa Breivika , da ima še " dve celici " , s katerima sodeluje , poroča BBC . Navedbe je podal med prvim zaslišanjem po bombnem napadu v Oslu , na katerem je tudi priznal napad .}\\
\midrule
golden & Policija njegove navedbe o sodelavcih še preiskuje\\
mT5-large &      Norveška policija preiskuje navedbe Andersa Behringa Breivika, da ima še " dve celici ", s katerima sodeluje \\
mBART-50-large &  Navedbe je podal med prvim zaslišanjem po bombnem napadu v Oslu \\
T5-sl-small\textsubscript{5} & Norveški policisti preiskujejo navedbe Andersa Behringa Breivika, da ima še dve celici, s katerima sodeluje\\
T5-sl-large\textsubscript{3} &  V prvem zaslišanju po bombnem napadu v Oslu je Breivik priznal napad \\
\bottomrule
\end{tabular}
\end{table}

\begin{table}
\caption{Translated examples of generated summaries on the ASN summarization task for the best performing models.}
\label{tab:asn-examples-en}
\begin{tabular}{lp{0.8\textwidth}}
\multicolumn{2}{p{\textwidth}}{Candidates will be notified of the selection by 3 April at the latest. As the main request, the Fund received 60 applications for apartments in Novo mesto, 53 for Postojna, 36 for Maribor, 43 for Kranj and 15 applications for apartments in Jesenice. Demand was most pronounced for locations in Postojna and Novo mesto, where the Fund is appearing for the first time and the response was therefore expected. The Fund will examine the received applications in detail, and the selection of buyers will take place on 27 March in the presence of a notary and a commission, in accordance with the provisions
The apartments that will remain unsold as part of the selection will be published by the Fund on its website www.ssrs.si on 3 April and sold under the conditions of the tender from 7 April onwards. All interested parties, not only those from specifically listed priority categories, can apply as buyers of apartments for resale.}\\
\midrule
golden &  The Housing Fund of the Republic of Slovenia received 207 applications in the deadline for submitting applications for the purchase of 364 market apartments in Novo mesto, Postojna, Maribor, Kranj and Jesenice, which expired on Monday.\\
mT5-large &       The Housing Fund in Slovenia has launched a tender for the sale of apartments for resale.\\
mBART-50-large &        The Krško Nuclear Power Plant (JEK) Fund for the Financing of Decommissioning has published a tender for the sale of apartments for resale.\\
T5-sl-small\textsubscript{5} &  Ljubljana - The Housing Fund (Fund) today published a tender for the purchase of apartments in Ljubljana, which was published by the Fund in early March.\\
T5-sl-large\textsubscript{3} &  The Housing Fund of the Republic of Slovenia has launched a tender for the sale of apartments for resale of apartments. \\
\midrule
\multicolumn{2}{p{\textwidth}}{Ivan Zidar of SCT, first lady of Vegrad Hilda Tovšak and the champion of Primorje Dušan Črnigoj can still be detained. The first two were released yesterday by the investigating judge and requested that a panel of three judges of the District Court in Ljubljana decide on the ordering of detention. Črnigoj was released on the grounds that he had invoked immunity. The prosecutor's office has already sent a request to the National Council today to revoke the suspect's immunity. Zidar and the Tovšak were too generous 
Criminal charges against Zidar, Tovšak and three others accused of committing the criminal offence of illicit giving of gifts under Article 248 (1) of the Criminal Code, which says: "Whoever promises, offers or gives to a person engaged in an economic activity an unauthorised prize, a gift or some other benefit for it or for someone else in order to obtain for himself or someone else any unjustified advantage in the conclusion of a transaction or performance referred to in the first paragraph of Article 247 of this Code (unauthorised acceptance of gifts) shall be punishable by imprisonment of six months to five years."
Only Tomaž Žibert was detained
At today's press conference, Harij Furlan, head of the organized crime prosecution team at the Supreme Prosecutor's Office, said that seven persons accused of giving and receiving gifts had been detained.
The investigating judge ordered only a member of the tender commission of the public company Air Traffic Control Tomaž Žibert to be detained.}\\
\midrule
golden & Ivan Zidar and Hilda Tovšak, suspected of illicit gift - giving, may still be placed in custody if a panel of three judges so decides. Črnigoj must first be stripped of his immunity.\\
mT5-large &      The investigating judge yesterday released Ivan Zidar, Hilda Tovšak and Dušan Črnigoj \\
mBART-50-large & 
The prosecution has already sent a request to the National Council to waive the suspect's immunity. \\
T5-sl-small\textsubscript{5} & Ivan Zidar, Hilda Tovšak and Dušan Črnigoj can still be placed in custody.\\
T5-sl-large\textsubscript{3} & Ivan Zidar, Hilda Tovšak and Dušan Črnigoj can still be placed in custody\\
\midrule
\multicolumn{2}{p{\textwidth}}{ Norwegian police are investigating allegations made by Anders Behring Breivik that he has "two more cells" he is cooperating with, reports the BBC.
He made the allegations during the first hearing after the bombing in Oslo, when he also admitted to the attack.}\\
\midrule
golden & Police are still investigating his allegations of coworkers\\
mT5-large &  Norwegian police are investigating Anders Behring Breivik's claim that he has "two more cells" with which he is cooperating \\
mBART-50-large &  He made the allegations during the first hearing after the bombing in Oslo \\
T5-sl-small\textsubscript{5} & Norwegian policemen are investigating allegations made by Anders Behring Breivik that he has two more cells with which he is cooperating \\
T5-sl-large\textsubscript{3} &  In first hearing after Oslo bombing, Breivik admitted to the attack \\
\bottomrule
\end{tabular}
\end{table}

\end{document}